\RequirePackage{fix-cm}
\documentclass[onecolumn]{svjour3}       
\smartqed  
\usepackage{graphicx}

\usepackage{amsmath}
\usepackage{graphicx}
\usepackage{setspace}
\usepackage{tocloft}
\usepackage{mathrsfs}  
\usepackage{hyperref}
\hypersetup{
    colorlinks=true,
    linkcolor=blue,
    filecolor=magenta,      
    urlcolor=cyan,
}
\begin{document}

\title{Can We See Photosynthesis?
}
\subtitle{ Magnifying the Tiny Color Changes of Plant Green Leaves Using Eulerian Video Magnification}

\titlerunning{Can We See Photosynthesis?}        

\author{Islam A.T.F. Taj-Eddin      \and
        Mahmoud Afifi \and
        Mostafa Korashy \and
        Ali H. Ahmed \and
		Ng Yoke Cheng \and
        Evelyng Hernandez \and
        Salma M. Abdel-latif
}

\authorrunning{Islam Taj-Eddin et al.} 

\institute{Islam A.T.F. Taj-Eddin \at
             Assiut University, Information Technology Dept., Egypt\\
              \email{Itajeddin@fci.au.edu.eg}           
           \and
            Mahmoud Afifi \at
              York University, Electrical Engineering and Computer Science Dept., Canada
              \and 
              Mostafa Korashy \at
              Assiut University, Information Technology Dept., Egypt 
              \and
              Ali H.Ahmed \at 
              Assiut University, Information Technology Dept., Egypt
              \and
               Ng Yoke Cheng \at
                           University of London, Singapore
             \and 
             Evelyng Hernandez
             \at San Francisco State University, USA  
             \and
                          Salma M. Abdel-latif \at
                          Assiut University, Botany and Microbiology Dept., Egypt 
}

\maketitle

\begin{abstract}
Plant aliveness is proven through laboratory experiments and special scientific instruments. In this paper, we aim to detect the degree of animation of plants based on the magnification of the small color changes in the plant's green leaves using the Eulerian video magnification. Capturing the video under a controlled environment, e.g., using a tripod and direct current (DC) light sources, reduces camera movements and minimizes light fluctuations; we aim to reduce the external factors as much as possible. The acquired video is then stabilized and a proposed algorithm used to reduce the illumination variations. Lastly, the Euler magnification is utilized to magnify the color changes on the light invariant video. The proposed system does not require any special purpose instruments as it uses a digital camera with a regular frame rate. The results of magnified color changes on both natural and plastic leaves show that the live green leaves have color changes in contrast to the plastic leaves. Hence, we can argue that the color changes of the leaves are due to biological operations, such as photosynthesis. To date, this is possibly the first work that focuses on interpreting visually, some biological operations of plants without any special purpose instruments. 
\end{abstract}

\section{Introduction}
\label{intro}
\begin{figure}
\centering
 \includegraphics[width=\linewidth]{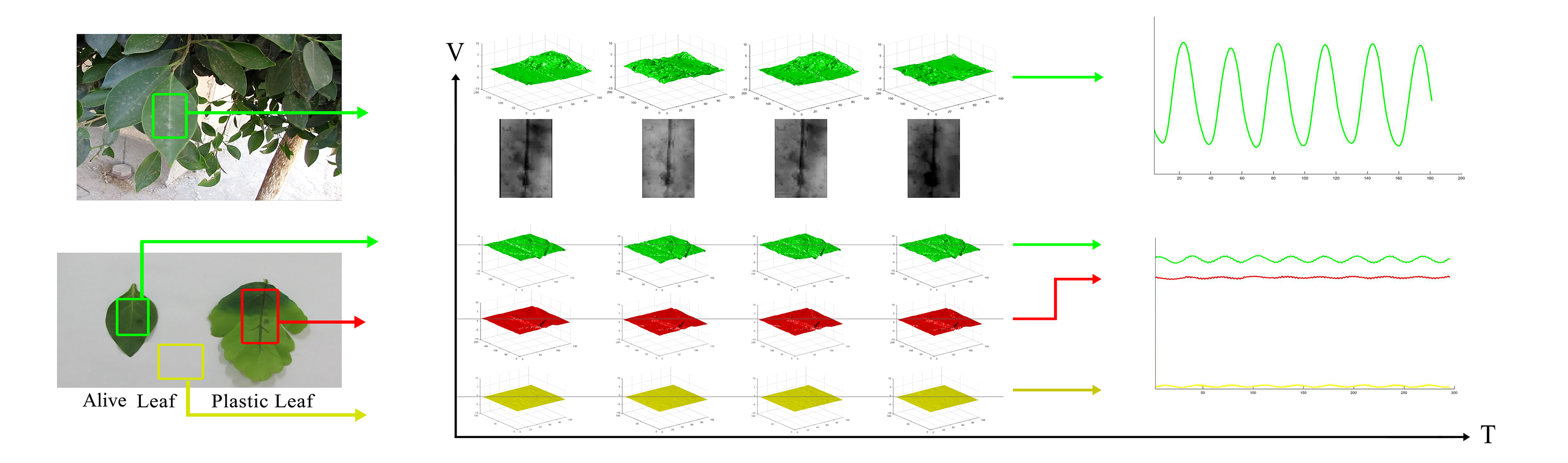}
   \caption{Upper photo: The result of magnifying the subtle changes of a live green leaf demonstrates large changes in the leaf color. Lower photo: The result of magnifying the subtle changes in the color of a plastic green leaf demonstrates small changes in the leaf color that is similar to the changes shown in the background. The noticeable difference between a live green leaf compared with the background and the plastic green leaf is shown. See the supplementary \href{https://www.dropbox.com/s/op64357dlnms4uq/Can 2020We 2020See 2020Photosynthesis__.mp4?dl=0g}{video}.}
    \label{fig:result1}
\end{figure}

Chlorophyll is a term used for green pigments found in plants. It has an important role in the photosynthesis process. When a molecule of chlorophyll absorbs light, it is promoted from its ground state to the excited state. The molecule in the excited state contains energy that is either: 
\begin{enumerate}
\item Passed to another chlorophyll molecule where energy is used in a two-step photosynthesis, photosystem I and photosystem II (i.e. photochemical quenching)  
\item Returned to the ground state by emitting the energy as heat dissipation (i.e.non-photochemical quenching) 
\item Returned to the ground state by emitting a photon (i.e. fluorescence)  
\end{enumerate}
Light absorption by photosynthetic pigments is extremely fast. The transition from the electronic ground state to a hyperactive state and the decays of the overactive state to the first singlet excited state occur within femtoseconds ($10^{-13} - 10^{-15}$ seconds) \cite{rabinowitch1969photosynthesis}. To use measurements of chlorophyll fluorescence in analyzing photosynthesis, researchers must distinguish between photochemical quenching and non-photochemical quenching. This is achieved by stopping photochemistry (i.e., photochemical quenching), which allows researchers to measure fluorescence with the sole presence of non-photochemical quenching. To reduce photochemical quenching to negligible levels, a high-intensity, short flash of light is applied to the green leaf. The high-intensity short flash can be produced mostly by laser beams or other means. This is the method of measuring fluorescence in natural or dead leaves. Existing technologies and methodologies use special instruments, such as Fluorimeter, to measure fluorescence \cite{cendrero2015dynamic}.  
\\
Eulerian Video Magnification is a recent technique for magnifying the tiny changes in colors that are not observable by the naked eyes\cite{Wu:2012:EVM:2185520.2185561}. This technique can be used to magnify the changes in the colors of a plant leaf to visualize the effect of chlorophyll fluorescence to the leaf colors. Because this technique requires no additional instruments and more cheap than the existing methodologies, it may replace the existing techniques and special purpose instruments. Such technique can help researchers and farmers in monitoring the status of plants without the need for bringing a sample to a laboratory and without any cost except a traditional camera.  
\\
The main contribution of this work is to visualize, without using special purpose instruments, the tiny changes of colors on the green leaf of the plant. The tiny changes in color observed are the by-product of the excited state process (i.e., photochemical quenching, non-photochemical quenching and fluorescence imitation). Chlorophyll fluorescence imaging is a useful non-destructive method to monitor the health status of plants. It shows how the color changes of live plants differ from the small changes of inanimate objects, see Fig. \ref{fig:result1}.
\\
The rest of the paper is organized as follows. In Section \ref{ourMethod}, we present in details the proposed system. The experimental results are presented in Section \ref{results}, followed by the conclusion in Section \ref{conclusion}.


\section{Methodology}
\label{ourMethod}
The proposed system aims to use a regular digital camera to visualize the changes of the live leaf's colors that represent the Chlorophyll fluorescence without using any special purpose instruments. In this section, we present two main steps that are used to overcome the external factors, namely stabilizing and reduction of illumination variation. Lastly, we present the color magnification step. 
\subsection{Stabilizing}
The objective of this stage is to eliminate the remaining undesirable shakes or movements of both camera and plant leaves. The corner points of the reference object are extracted using minimum eigenvalue algorithm \cite{eigenvalue}, then the Kanade-Lucas-Tomasi (KLT) algorithm \cite{tomasi1991detection} is utilized using forward-backward error thresholding \cite{bidirectional}. For each frame, the transformation matrix between the corner points of the current frame and the corner points of the previous frame is estimated using M-estimator sample consensus algorithm \cite{MSAC}. Consequently, we use the inverse of the transformation matrix to reduce the movements of the reference object.
\\
The recorded video is stabilized in two phases. The first phase is to reduce the camera shaking, while the second phase is to eliminate undesirable movements of the Region of Interest (ROI) of the plant. For the first phase of stabilization, a static rigid body at the background is determined by the experimenter. This rigid body is used as a reference object in order to eliminate camera vibrations. 
For the second phase, in order to generate a video of an immobilized green leaf, the experimenter re-applies the stabilization process to the (ROI) of the green leaf that is determined by the user as well.
\subsection{Reduction of Illumination Variations}
We minimize the changes of the illumination to emphasize that color changes are due to changes in the plant leaf color and not caused by any other reason. The background is separated by removing the leaves from the input video. We use the traditional chroma keying technique to exclude green pixels $Pg$ and generate a video without green pixels $V{-P_g}$. Then, we generate an approximate illumination invariant video (normalized video). Given an input video, we generate a video $V{-P_g}$ without green pixels. For each $n \times m$ frame in $V{-P_g}$, we construct a 1D vector (1$\times (n \times m$)) $I$ that represents the 1D representation of the frame. A normalized 1D vector $I_{norm}$ is generated using homomorphic filtering-based normalization technique \cite{HOMO}.
\\
For each color frame $I_{norm_{i}}$, we calculate a vector $\vec{v_i}$ that represents the changes of lighting with time as shown in the following equations
\begin{equation}
\label{e1}
\vec{v}=\langle M(I_{norm_{1}}),M(I_{norm_{2}}),...,M(I_{norm_{L-1}})\rangle,
\end{equation}
\begin{equation}
M(I_{norm_{i}})=\frac{\rVert{I_{norm_{i}}-I_{norm_{i-1}}}\lVert_1}{(n \times m)},
\end{equation}
\noindent
where $M(I_{norm_{i}})$ is a kernel function that calculates the mean of the first order derivative of  $I_{norm_{i}}$ w.r.t. time, $i$ is the number of the current frame in the video, and $L$ represents the number of frames in the given video.
We generate a signal $S$ that represents the normalized video, i.e., an estimated lighting invariant version, using the following equation
\begin{equation}
\label{e3}
S_i=\langle F_i-\vec{v}_{(i)}\rangle,
\end{equation}
 where $F_i$ is the frame number $i$ of the original video.
\\
Although we attempt to reduce both lighting changes and camera movements in a reasonable manner, high illumination variations, e.g. outdoor lighting, and camera movements lead to possibly inaccurate results. Thus, we suggest conducting the experiments under controlled environment (i.e., indoor lighting conditions and a semi-static or static camera).
\subsection{Color Magnification} To magnify the changes of the (ROI) colors, we use Euler magnification technique \cite{Wu:2012:EVM:2185520.2185561}.
The given video is decomposed into different spatial frequency bands using a spatial pyramid. Temporal filtering is applied to all bands to extract the frequency band of interest. The magnified video is reconstructed using the amplified signals of those frequencies which are added back to the original signal. The selected band of temporal frequencies varies based on the case. According to the extremely high rate of the fluorescence, we use large values of the amplification factor $\alpha$ and the spatial wavelength $\lambda$ to magnify the changes of colors as much as possible. In addition, we cut off the amplification factor of the blue channel to reduce the generated noise; where the blue channel 
is the color channel that is most susceptible to the noise \cite{multiresolution}.


\section{Experimental Results}
\label{results}

\begin{figure*}[t!]
  \centering
 \includegraphics[width=\linewidth]{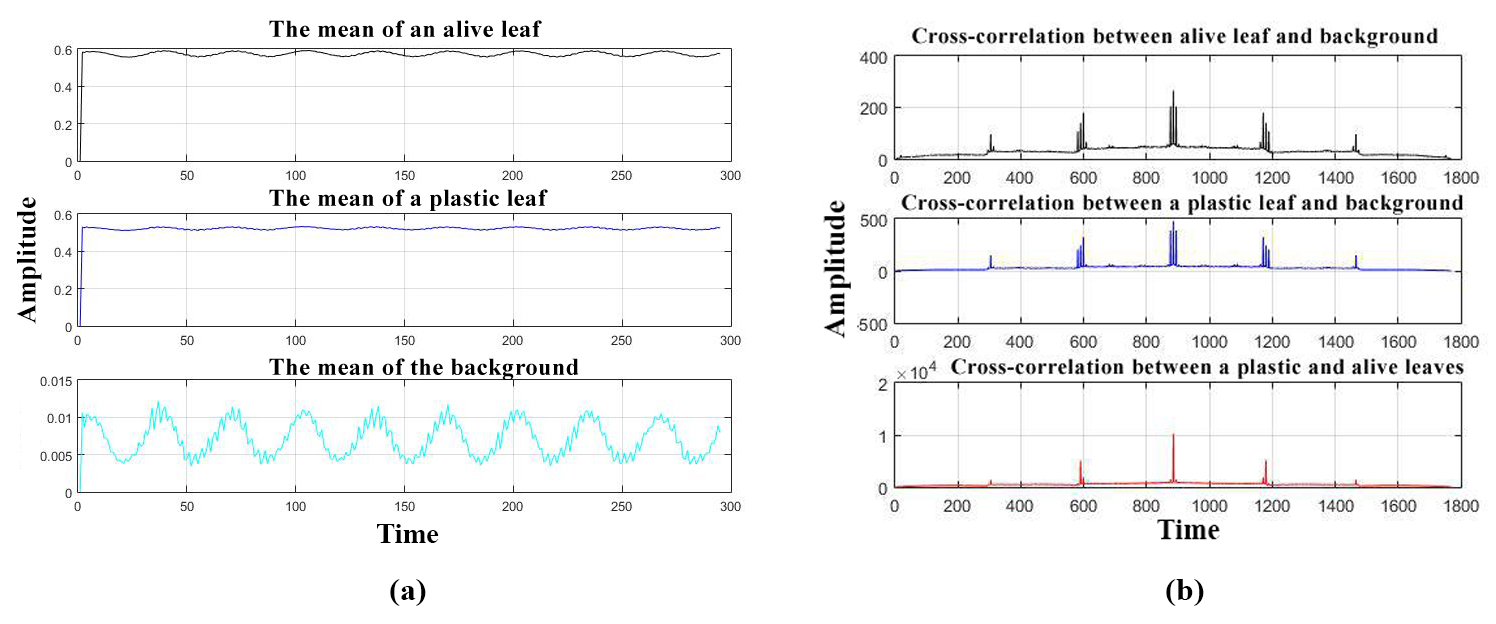}

  \caption{(a) The mean of live green leaf, plastic leaf, and background. (b) The similarity between background and live green leaf, background and plastic green leaf, live green leaf and plastic green leaf.}
  \label{fig:result2}
 
\end{figure*}

\begin{figure}
\centering
 \includegraphics[width=1\linewidth]{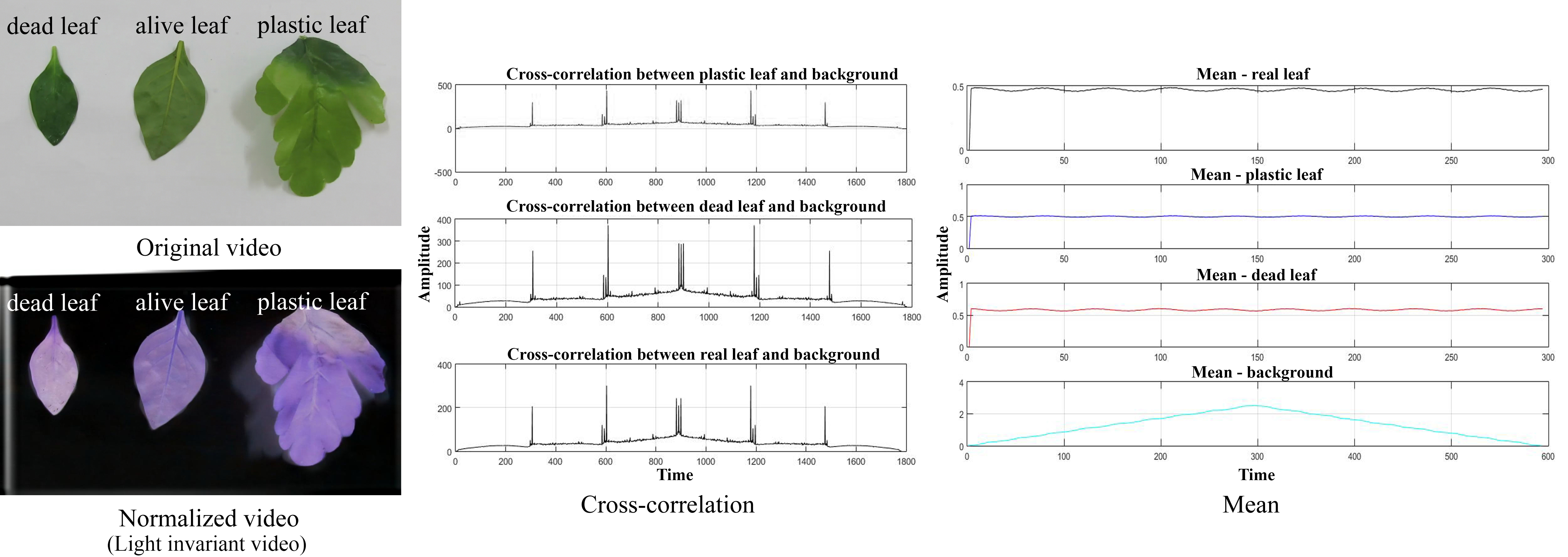}
   \caption{A comparison among color changes over time on a live, a dead, and a plastic leaves.
   }
    \label{fig:result4}
\end{figure}

We have applied the proposed system to HD video sequences at 30 frames per second (fps). We have used $\alpha=120$ and $\lambda=1000$ for magnifying the color changes. The results show that the pixel intensities of the (ROI) for the live plant oscillate in a manner worth noting, compared to an inanimate object.
\\
Fig. \ref{fig:result1} shows two case studies.  The upper part of the figure represents a live green leaf that was captured under outdoor lighting conditions. The middle part shows the 3D visualization of the color vibrations of the (ROI). The left part represents the frequency of color changes with time. The lower part of the figure shows the result of magnifying the color changes of a plastic green leaf, a live plant green leaf, and a segment of the background. As shown, there are small changes in the both plastic green leaf color and the background. However, the live green leaf has noticeable color changes as shown in the 3D visualization and frequency of the magnified changes of colors. Fig. \ref{fig:result2}-a enlarges the mean of color changes of a segment of the background, a plastic green leaf, and a live green leaf that were used in another experiment under controlled indoor lighting conditions.
\\\\
\textbf{Proof of Concept:} We have recorded a video for a live green leaf, and a plastic green leaf under the same indoor lighting conditions. We study the frequency cross-correlation of the following:

\begin{enumerate}
\item A patch of the live green leaf and the background
\item A patch of the plastic green leaf and the background
\item A patch of the live and plastic leaves
\end{enumerate}
As shown in Fig. \ref{fig:result2}-b, the cross-correlation between the live and plastic leaves has the maximum amplitude because both of them have similar green color. This similarity of colors has its effect on reflecting the incoming light to be captured in the video. Nevertheless, the cross-correlation between the live green leaf and the background is smaller than the cross-correlation between the plastic green leaf and the background. As shown in Figure \ref{fig:result4}, another experiment was performed using a dead green leaf (boiled at 100 Degree Celsius for 10 minutes) and a live green leaf.  As illustrated in the figure, the cross-correlation between the color changes on the live green leaf and a segment of the background is smaller than the cross-correlation between the color changes on the dead green leaf and the patch of the background. This indicates that the live green leaf undergoes noticeable tiny color changes compared to the plastic green leaf. As aforementioned, that tiny changes in color is the by-product of the excited state process (i.e., photochemical quenching, non-photochemical quenching and fluorescence imitation) that happen due to some green leaf physiological activities (i.e., photosynthesis) that makes the color of its leaves oscillate as shown in Fig. \ref{fig:result1}. 

\subsection{Comparisons}
In order to get a quantitative comparison between the color changes of the live and dead leaves, we have recorded 32 videos for eight leaves in two cases: alive and dead, under both DC and AC lighting \footnote{The videos are available in the following link: \href{https://goo.gl/ezEPzn}{https://goo.gl/ezEPzn}}. Then, we have applied the proposed system to get 2,500 samples of the magnified colors of the ROI for each leaf (total number of samples are 40,000 samples per lighting condition). Each sample is represented by the first order derivative of the magnified color w.r.t. time of a single pixel in the ROI. Finally, we have trained a radial basis function SVM classifier using these samples to get an approximated discrimination level of the color changes of the live and dead leaves. In the end, we have reported the mean accuracy of the 5-fold cross validation. The classification accuracy is 64.28\% and 62.57\% using videos illuminated by a DC light source and a regular light source (AC), respectively. Consequently, we can say that the color changes of the live leaves have something different, compared with dead leaves. 
\section{Conclusion}
\label{conclusion}
In this work, we have presented a system to visualize the subtle color changes of a live plant green leaf, which is a by-product of the photosynthesis process, without using any special purpose instruments. Firstly, the acquired video is stabilized to eliminate, as much as possible, undesirable movements of both plant leaf samples and camera. The illumination variations are reduced to emphasize the color changes that are produced only by the plant leaves. The lighting invariant version of the stabilized video is used to magnify the subtle changes of colors in the region of interest using Eulerian video magnification technique. The mean of color changes of the live plant leaf differs from that of inanimate objects. We have demonstrated that by using a regular digital camera, the aliveness of green leaves could be detected by studying the frequencies of color changes which is due to biological operations (i.e., photosynthesis).



\begin{thebibliography}{}
\bibitem{cendrero2015dynamic}
Cendrero-Mateo M. Pilar, Carmo-Silva A. Elizabete, Porcar-Castell Albert, Hamerlynck Erik P., Papuga Shirley A., and Moran M. Susan \emph{Dynamic response of plant chlorophyll fluorescence to light, water and nutrient availability}, Functional Plant Biology 42, 746–-757, 2015.

\bibitem{rabinowitch1969photosynthesis}
Rabinowitch, Eugene, \emph{Govindjee, Photosynthesis},  New York-London-Sydney-Toronto, 1969.


\bibitem{Daniel}
Daniel J McDuff, Justin R Estepp, Alyssa M Piasecki, and Ethan B Blackford, \emph{A survey
of remote optical photoplethysmographic imaging methods}, In 37th Annual International Conference of the IEEE Engineering in Medicine and Biology Society (EMBC), 6398-–6404. IEEE, 2015.


\bibitem{eigenvalue}
Shi, Jianbo, and Carlo Tomasi, \emph{Good features to track}, Proceedings of IEEE Computer Society Conference on Computer Vision and Pattern Recognition (CVPR'94), 593--600, 1994.


\bibitem{tomasi1991detection}
Tomasi, Carlo, and Takeo Kanade, \emph{Detection and tracking of point features}, Pittsburgh: School of Computer Science, Carnegie Mellon Univ., 1991.

\bibitem{bidirectional}
Zdenek Kalal, Krystian Mikolajczyk, and Jiri Matas, \emph{Forward-Backward Error: Automatic Detection of Tracking Failures}, In Proceedings of the 2010 20th International Conference on Pattern Recognition (ICPR '10). IEEE Computer Society, 2756--2759, 2010.

\bibitem{MSAC}
Torr, Philip HS, and David W. Murray, \emph{The development and comparison of robust methods for estimating the fundamental matrix}, International journal of computer vision 24.3: 271--300, 1997.
\bibitem{HOMO}
Heusch, Guillaume, Fabien Cardinaux, and Sébastien Marcel, \emph{Lighting normalization algorithms for face verification}, No. EPFL-REPORT-83268. IDIAP, 2005.

\bibitem{Wu:2012:EVM:2185520.2185561}
Wu, Hao-Yu, Michael Rubinstein, Eugene Shih, John Guttag, Frédo Durand, and William Freeman, \emph{Eulerian Video Magnification for Revealing Subtle Changes in the World}, ACM Transactions on Graphics 31, no. 4: 1–-8, 2012.
\bibitem{multiresolution}
Zhang, M. and Gunturk, B.K.,\emph{Multiresolution bilateral filtering for image denoising}, IEEE Transactions on image processing, 17(12): 2324--2333, 2008.
%
%
%
\end{thebibliography}


\end{document}